\documentclass[letterpaper]{article} % DO NOT CHANGE THIS
\usepackage{aaai24}  % DO NOT CHANGE THIS
\usepackage{times}  % DO NOT CHANGE THIS
\usepackage{helvet}  % DO NOT CHANGE THIS
\usepackage{courier}  % DO NOT CHANGE THIS
\usepackage[hyphens]{url}  % DO NOT CHANGE THIS
\usepackage{graphicx} % DO NOT CHANGE THIS
\usepackage{natbib}  % DO NOT CHANGE THIS AND DO NOT ADD ANY OPTIONS TO IT
\usepackage{caption} % DO NOT CHANGE THIS AND DO NOT ADD ANY OPTIONS TO IT
\usepackage{algorithm}
\usepackage{algorithmic}

\usepackage{newfloat}
\usepackage{listings}
\DeclareCaptionStyle{ruled}{labelfont=normalfont,labelsep=colon,strut=off} % DO NOT CHANGE THIS
\floatstyle{ruled}
\newfloat{listing}{tb}{lst}{}
\floatname{listing}{Listing}
\usepackage{tabularx}
\usepackage{multicol}
\usepackage{multirow}
\usepackage{todonotes}
\usepackage{amsmath,amsfonts}

\usepackage{xcolor}
\usepackage{makecell}

\newcommand{\n}{{{\textbackslash n}}}
\newcommand{\taskdesc}{{\{TASK DESC\}}}
\newcommand{\obs}{{ $\text{\{OBS\}}$}}

\newcommand{\invstate}{{ $\text{\{INV STATE\}}$}}
\newcommand{\score}{{$\text{\{SCORE\}}$}}
\newcommand{\validAction}{{ $\text{\{VALID ACT SET\}}$}}

\title{Large Language Models Are Neurosymbolic Reasoners}
\author{
    Meng Fang\equalcontrib\textsuperscript{\rm 1,\rm 2}, Shilong Deng\equalcontrib\textsuperscript{\rm 1}, Yudi Zhang\equalcontrib\textsuperscript{\rm 2}, \\
    Zijing Shi\textsuperscript{\rm 3}, Ling Chen\textsuperscript{\rm 3}, Mykola Pechenizkiy\textsuperscript{\rm 2}, Jun Wang\textsuperscript{\rm 4}
}
\affiliations{
    \textsuperscript{\rm 1}University of Liverpool, United Kingdom \\ \textsuperscript{\rm 2}Eindhoven University of Technology, Netherlands \\ 
    \textsuperscript{\rm 3}University of Technology Sydney, Australia \\ 
    \textsuperscript{\rm 4}University College London, United Kingdom \\
    \{Meng.Fang, shilong.deng\}@liverpool.ac.uk, \{y.zhang5, m.pechenizkiy\}@tue.nl, \\zijing.shi@student.uts.edu.au, ling.chen@uts.edu.au, j.wang@cs.ucl.ac.uk
}

\usepackage{bibentry}
\begin{document}

\maketitle

\begin{abstract}
%A wide range of real-world applications are characterized by their symbolic nature, necessitating a strong capability for symbolic reasoning and long-term planning. This paper investigates the potential application of Large Language Models (LLMs) as symbolic reasoners. We focus on text-based games, which are important benchmarks for agents with natural language capabilities, especially in their symbolic tasks, such as math, map reading, sorting, and text world common sense. To facilitate such agents, we propose an LLM-based agent designed to solve symbolic challenges and achieve in-game objectives. We begin by initializing the LLM-based agent, and informing it of its role. The agent then receives observations and a set of valid actions provided by the text-based games, alongside a symbolic module. With these inputs, the LLM-based agent is prompted to choose an action and interact with the game environments. Our experimental results demonstrate that our method significantly enhances the capability of LLMs as automated agents for symbolic reasoning. Our agent outperforms most baseline methods, including the Deep Reinforcement Relevance Network (DRRN) utilizing symbolic modules and the Behavior Cloned Transformer trained with abundant expert data, and achieves an average performance of 88\% across all tasks. 
A wide range of real-world applications is characterized by their symbolic nature, necessitating a strong capability for symbolic reasoning. This paper investigates the potential application of Large Language Models (LLMs) as symbolic reasoners. We focus on text-based games, significant benchmarks for agents with natural language capabilities, particularly in symbolic tasks like math, map reading, sorting, and applying common sense in text-based worlds. To facilitate these agents, we propose an LLM agent designed to tackle symbolic challenges and achieve in-game objectives. We begin by initializing the LLM agent and informing it of its role. The agent then receives observations and a set of valid actions from the text-based games, along with a specific symbolic module. With these inputs, the LLM agent chooses an action and interacts with the game environments. Our experimental results demonstrate that our method significantly enhances the capability of LLMs as automated agents for symbolic reasoning, and our LLM agent is effective in text-based games involving symbolic tasks, achieving an average performance of 88\% across all tasks.
\end{abstract}

\section{Introduction}
The ability to perform reasoning is crucial for AI due to its significant impact on various real-world tasks. The widespread adoption of large language models (LLMs), such as ChatGPT and GPT-4~\cite{OpenAI_GPT4_2023}, has led to a series of remarkable successes in reasoning tasks, ranging from question \& answering to solving math problems. Among these challenges, text-based games serve as important benchmarks for agents with natural language capabilities and have garnered significant attention 
%in the realm of language-centric machine learning research~\cite{narasimhan2015language, cote2019textworld, yao-etal-2020-keep, chaudhury2021neuro, osborne2022survey}. 
in the realm of language-centric machine learning research~\cite{narasimhan2015language, cote2019textworld, xu2020deep, ryu-etal-2022-fire, shi2022stay}. 
In these games, an agent uses language to interpret various scenarios and make decisions. The complexity of such games arises from the need for language comprehension, common sense, managing action spaces with combinatorial complexity, and the crucial importance of long-term memory and planning~\cite{cote2019textworld,Wang2022ScienceWorld}. The challenges escalate in text-based games that 
%involve symbolic tasks~\cite{lample2019deep, poesia2021contrastive, bc_symbolic, qian-etal-2023-limitations}. 
involve symbolic tasks~\cite{bc_symbolic}. 
For instance, contemporary agents might be tasked with a scenario where they are required to solve a mathematical problem and simultaneously gather a specified amount of fruits, with the quantity needed being the solution to the math problem.

Using symbolic modules or external tools for arithmetic, navigation, sorting, and knowledge-base lookup is crucial for language agents, especially in complex text-based games~\cite{lample2019deep, poesia2021contrastive, bc_symbolic, qian-etal-2023-limitations}. However, effectively integrating these aspects into language agents remains a relatively unaddressed challenge. Solving such text-based games requires interactive multi-step reasoning, and agents have most commonly been modeled using reinforcement learning~\cite{xu2020deep,yao-etal-2020-keep,xu-etal-2021-generalization-text}. These methods, however, face challenges such as delayed rewards and difficulty in exploring large action spaces. Recently, there has been an exploration of imitation learning approaches, which utilize human play data~\cite{bc_symbolic}. While Behavior Cloning (BC) shows potential in effectively addressing these challenges, it often necessitates substantial effort and resources. This is primarily due to the need for acquiring expert data.

%Recently, large language models (LLMs) have demonstrated in-context generalization capabilities, and it shows the possibility of eliciting reasoning abilities by prompting LLMs \cite{brown2020language,min-etal-2022-metaicl}. However, it is relatively unexplored that LLMs can be prompted to perform symbolic reasoning. These large models, such as GPT-3.5 and GPT-4, exhibit the capability to encode extensive amounts of information \cite{OpenAI_GPT4_2023}. One notable example is the acquisition of substantial knowledge by LLMs through its training process, resulting in their ability to approach human-level performance when addressing an extensive variety of tasks~\cite{OpenAI_GPT4_2023}. It seems that we can directly leverage LLMs as a neurosymoblic reasoner without labeled gold training data. However, there is little work reasoning with logic, graphs, or symbolic formulas. It is very interesting to investigate the development of a method based on LLM for the purpose of symbolic reasoning.
Recently, large language models (LLMs) have demonstrated notable in-context generalization capabilities, suggesting the potential to elicit reasoning abilities by prompting these models~\cite{brown2020language,min2021metaicl}. However, the application of LLMs in performing symbolic reasoning remains an under-explored area. Models like GPT-3.5 and GPT-4 have shown the ability to encode extensive information~\cite{OpenAI_GPT4_2023}. A significant example of this is their acquisition of substantial knowledge during training, enabling them to approach human-level performance across a wide range of tasks~\cite{OpenAI_GPT4_2023}. This indicates the feasibility of utilizing LLMs as neurosymbolic reasoners without relying on labeled gold training data. However, there is currently limited research on utilizing these models for reasoning tasks that involve logic, graphs, or symbolic formulas. The exploration and development of methods that leverage LLMs for symbolic reasoning is highly intriguing and holds significant potential impact.
%However, researchers have come across limitations in the field of symbolic reasoning on LLMs. These limitations pertain to the ability to modify and deduce structured representations, such as logic, graphs, or symbolic formulas. Representing the chain of thought of LLMs in symbolic logic presents a significant challenge. Thus it is very interesting to investigate the development of a method based on LLM for the purpose of symbolic reasoning.

\begin{figure}[t]
\centering
\includegraphics[width=0.85\columnwidth]{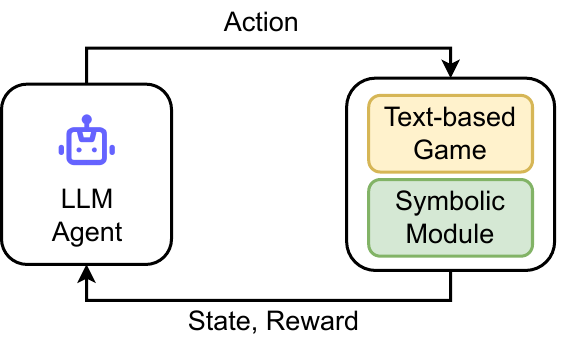}
\caption{The LLM agent is capable of interacting with the game environment, leveraging its reasoning abilities to determine the most suitable actions. These actions alter the environment's state and contribute to achieving the given objective. The environment, along with its corresponding symbolic modules, offers a valid set of actions to the LLM agent. The agent's responsibility is to select an action from this set. The chosen action will then dictate how the agent interacts with either the game environment or the symbolic module.}
\label{fig1}
\end{figure}

In this paper, our aim is to investigate the role of Large Language Models (LLMs) in symbolic reasoning within the context of text-based games. When engaging in games that involve symbolic tasks, our LLM agent generates the most rational actions based on the observed game state in a zero-shot manner, assisted by external symbolic modules such as calculators or navigators, as illustrated in Figure~\ref{fig1}. The LLM agent employs both the text-based game environment and symbolic modules to generate a list of valid actions. These valid actions, along with the current observation, are integrated into the prompt to direct the LLM agent in selecting an appropriate action. Subsequently, the LLM agent executes this action, interacting with both the game environment and symbolic modules to complete the task.

%We summarize our main contributions as follows:
In summary, our contributions include:
\begin{itemize}
  \item We introduce the use of LLMs for symbolic reasoning and provide a framework for employing the LLM agent as a neurosymbolic reasoner. This achievement underscores the potential of LLMs, with the support of external modules, to function as neurosymbolic reasoners, capable of successfully completing complex tasks.
    \item We have developed the LLM agent with tailored prompts, enabling it to effectively utilize symbolic modules and enhance its performance in text-based games that involve symbolic tasks.
%\item Thirdly, our experiments demonstrate that our LLM agent is effective in text-based games involving symbolic tasks, achieving an average performance of 88\% across all tasks. 
\item Our experiments demonstrate that our agent outperforms strong baselines, including the Deep Reinforcement Relevance Network with symbolic modules and the Behavior Cloned Transformer trained with extensive expert data, achieving an average performance of 88\% across all tasks.\footnote{Code at: https://github.com/hyintell/LLMSymbolic.} %. This approach outperforms strong baselines, such as the Deep Reinforcement Relevance Network (DRRN) and Behaviour Cloned (BC) Transformer. 
    %Although the performance of our model is slightly lower than that of the BC Transformer with symbolic modules.  it offers significant advantages in avoiding data collection and saving computational resources.
\end{itemize}

\section{Related Work}

%We investigate the application of LLMs as neurosymbolic reasoners, building upon text-based games that encompass symbolic tasks. Below we review related works in LLMs for decision making, neurosymbolic reasoning and text-based game playing agent.

\paragraph{Large Language Models for Decision Making.}
LLMs have demonstrated notable capabilities, enabling their application in tasks that extend beyond language generation \cite{OpenAI_GPT4_2023}. Furthermore, they are increasingly being grounded as policy models for decision-making in interactive contexts \cite{yang2023foundation}. Current studies focus on enhancing the decision-making capacity of LLMs through techniques such as prompting and in-context learning. For instance, \citet{weichain} introduce the Chain-of-Thought (CoT) approach, showing that a sequence of intermediate reasoning steps can enhance decision-making capabilities. \citet{yao2022react} present ReAct, a method for interleaved reasoning and action generation to improve performance in interactive decision-making tasks. Other studies \cite{singh2023progprompt, huang2022language,huang2022inner, liang2023code, vemprala2023chatgpt} have explored innovative strategies involving prompt engineering and the utilization of high-level function libraries to enhance the capabilities of LLMs. Additionally, some approaches incorporate mechanisms of self-critique and self-reflection into LLMs, enabling them to refine their generation. For example, \citet{shinn2023reflexion} introduce Reflexion, a technique that employs external feedback to detect ineffective actions and engage in self-reflection. \citet{madaan2023self} enable an LLM to offer feedback on its previously generated text and refine it adaptively. Recent attempts have also explored different aspects of LLMs for decision-making. \citet{kwon2023reward} utilize LLMs as proxy reward functions by prompting them with desired behaviors, while \citet{brooks2022context} consider LLMs as world models, where the agent learns policy by interacting with the LLM-based world model. In our work, we focus on developing suitable prompting strategies to enhance the decision-making performance of LLMs in solving symbolic tasks.

% Recent attempts explored different aspects of LLMs for decision-making. \citet{Ahn2022DoAI} use LLMs to generate plans or sub-goals that guide low-level Reinforcement Learning (RL) agents in taking actions. \citet{kwon2023reward} utilize LLMs as proxy reward functions by prompting them with desired behaviors. \citet{yao2022react} focus on enabling LLM agents to select actions in text-based environments. In addition, a recent approach considers LLMs as world models \cite{brooks2022context}, where the agent learns the policy by interacting with the LLM-based world model. 
% In this paper, we focus on directly grounding LLM in decision-making to take action to avoid learning an extra decision-making agent, which requires more data samples and time-consuming training.

\paragraph{Text-based Game.}
Text-based games can be formally characterized as partially observable Markov decision processes (POMDPs) \cite{cote2019textworld}. In recent years, there has been a notable increase in the design of reinforcement learning (RL) agents to solve these games \cite{liu2021learning,hendrycks2021would,osborne2022survey}. Current research primarily addresses challenges such as long-term dependencies, partial state observations, sparse rewards, and complex action combinations in text-based games \cite{yin2019comprehensible,ammanabrolu2020graph,kimura2021neuro,xu2022perceiving}. For instance, \citet{adhikari2020learning} address the challenge of partial observability by exploring the acquisition of graph-structured state representations through data-driven methods. \citet{yao-etal-2020-keep} and \citet{shi2022stay} employ a language model to generate a compact set of action candidates for RL agents, tackling the issue of the combinatorial action space. More recently, with the advancement of LLMs, research has shifted towards using prompts to enable LLMs to solve text-based games \cite{yao2022react,shinn2023reflexion}. However, these efforts have primarily focused on the LLMs' capability for in-context learning, while the exploration of their potential in symbolic reasoning has been relatively overlooked.

\paragraph{Neurosymbolic Reasoning.}
The field of neurosymbolic reasoning combines the capabilities of deep neural networks with symbolic reasoning, significantly reducing the search space associated with symbolic techniques. This approach has been used to tackle various complex multi-step inference challenges, including tasks like multi-hop question answering \cite{weber2019nlprolog}, language contextualization \cite{zellers2021piglet}, and semantic analysis \cite{cambria2022senticnet}. Text-based games that involve symbolic tasks serve as a valuable test-bed for addressing such challenges. Previous approaches have employed traditional optimization techniques or reinforcement learning agents. For example, \citet{kimura2021loa} decompose text-based games into collections of logical rules, which are then integrated with deep reinforcement learning. \citet{basu2022hybrid} use Integer Linear Programming (ILP) to substantially improve agent performance, providing an interpretable framework for understanding agents' selection of specific actions.
% This work aims to discuss the framework of text games and the methods proposed by \cite{kimura2021loa} to decompose text games into a collection of logical rules. Furthermore, the paper explores the integration of these rules with deep reinforcement learning, as suggested by \cite{kimura2021neuro} . The authors of the study conducted by \cite{basu2022hybrid} employed the use of Integer linear programming to significantly enhance the performance of agents. This approach also offered a more comprehensible framework for comprehending the rationale behind agents' decision-making process, as highlighted by \cite{chaudhury2021neuro}.
% This paper will talk about the text games, \cite{kimura2021loa} develop methods to decompose text games into a set of logical rules, then combine these rules with deep reinforcement learning \cite{kimura2021neuro}. \citet{basu2022hybrid} utilized Integer linear programming to substantially increase agent performance while providing a more interpretable framework for understanding why agents choose specific actions \cite{chaudhury2021neuro}. 
% More generally, neurosymbolic reasoning has been applied to a variety of multi-step inference problems, such as multi-hop question answering \cite{weber2019nlprolog}, language grounding \cite{zellers2021piglet}, and semantic analysis \cite{cambria2022senticnet}.

\begin{table*}[t]
\small
    \centering
    % \begin{tabularx}{0.5\textwidth}{X|X}
    % \begin{tabularx}{1\textwidth}{>{\hsize=.4\hsize}X|>{\hsize=1\hsize}X|>{\hsize=1.2\hsize}X}
    \begin{tabularx}{1\linewidth}{m{2cm}|m{7.5cm}m{7cm}}
    \hline
    Task 
    
    (Symbolic 
    
    Module) & Description & Symbolic Module\\ \hline
    Arithmetic (Calculation Module) & Your first task is to solve the math problem. Then, pick up the item with the same quantity as the math problem answer, and place it in the box. & INPUT: mul 8 7 
    
    RESPONSE: Multiplying 8 and 7 results in 56. \\ \hline
    MapReader 
    
    (Navigation Module) & Your task is to take the coin located in the pantry, and put it into the box found in the chamber. A map is provided, that you may find helpful. & %\textbf{POBS:} You are in chamber. You see a box that is empty. To the north, you see canteen. 
    
    INPUT: next step to pantry 
    
    RESPONSE: The next location to go to is canteen. If you want to go to pantry from chamber, you need go through canteen, pantry. \\ \hline
    
    Sorting 
    
    (Sorting 
    
    Module) & Your task is to sort objects by quantity.  First, place the object with the smallest quantity in the box.  Then, place the objects with the next smallest quantity in the box, and repeat until all objects have been placed in the box. & INPUT: sort ascending
    
    RESPONSE: The observed items, sorted in order of increasing quantity, are: 25 g of oak, 47 g of brick, 15 kg of cedar, 21 kg of marble. \\ \hline
    TWC
    
    (Knowledge Base Module) &  Your task is to pick up objects, then place them in their usual locations in the environment. & INPUT: query clean brown shirt 
    
    RESPONSE: Clean brown shirt is expected to be located at wardrobe. \\ \hline
    
    \end{tabularx}
    \caption{Text-based games with symbolic tasks and their corresponding symbolic modules. INPUT refers to the current action that is sent to the symbolic modules. RESPONSE denotes the responses generated by the symbolic modules at the present time.
    }
    \label{tab:task&module}
    % \vspace{-0.4cm}
\end{table*}

\section{Preliminaries}

\paragraph{Text-based Games as POMDPs.}
Text-based games can be formally defined as partially observable Markov decision processes (POMDPs), considering that the agent only observes partial information about the environment at each turn \cite{sutton2018reinforcement}. 
%When these games include symbolic tasks, agents are encouraged to take advantage of external symbolic modules in order to solve the provided problems~\cite{bc_symbolic}. More precisely, 
In games with symbolic modules, at each discrete time step $t$, the agent is provided with an observation denoted as $o_t$ and is given a task description denoted as $d$. The symbolic module then produces a collection of valid actions, denoted as $A_{t, SyM}$, while the text game environment concurrently establishes its own set of proper actions, denoted as $A_{t, {Env}}$. Consequently, the set of acceptable actions at time step $t$ is the union of these two sets, denoted as $A_t = A_{t, Env} \cup A_{t, SyM}$. The agent's goal is to select an action $a_t$ from the set of valid actions $A_t$, given the observation $o_t$ and the task description $d$. If $a_t$ belongs to the set $A_{t, SyM}$, the symbolic module generates the next observation $o_{t+1}$. Conversely, if $a_t$ is not part of $A_{t, SyM}$, the text-based game environment processes $a_t$ and produces both the subsequent observation $o_{t+1}$ and the reward $r_t$.

% Text-based games can be formally depicted as partially observable Markov decision processes (POMDPs) \cite{sutton2018reinforcement, cassandra1998survey}. When these games incorporate symbolic tasks, agents are motivated to utilize external symbolic modules to resolve the given problems~\cite{bc_symbolic}. Specifically, at each time step $t$, the agent makes an observation $o_t$ and receives a task description $d$. Following this, the symbolic module generates a set of valid actions, $A_{t, \text{SyM}}$, while the text game environment simultaneously establishes its unique set of valid actions, $A_{t, \text{Env}}$. Consequently, the set of valid actions at step $t$ is the union of these two sets, i.e., $A_t = A_{t, \text{Env}} \cup A_{t, \text{SyM}}$. Given $o_t$, task description $d$, and the valid action set $A_t$, the agent is tasked with selecting an action $a_t \in A_t$. If $a_t \in A_{t, \text{SyM}}$, the symbolic module generates the subsequent observation $o_{t+1}$. However, if $a_t$ is not part of $A_{t, \text{Sym}}$, the text game environment steps in, handling $a_t$ and producing both the next observation $o_{t+1}$ and the reward $r_t$.

\paragraph{Symbolic Tasks.}
%This study examines four distinct games within text-based games, namely Text World Common Sense (TWC), MapReader, Arithmetic, and Sorting. These tasks are explored within the setting as in the previous research \cite{bc_symbolic}. Each task is equipped with its own symbolic modules designed to assist LLMs in successfully accomplishing the task. The specific details pertaining to the four tasks are presented in the subsequent paragraph, while Table \ref{tab:task&module} provides concrete examples for each task.
There are four distinct tasks within text-based games, namely Arithmetic, MapReader, Sorting and Text World Common Sense (TWC)~\cite{bc_symbolic}. Each task is equipped with its own symbolic modules designed to assist agents in successfully accomplishing the task. 

\begin{figure}[t]
\centering
\includegraphics[width=\linewidth]{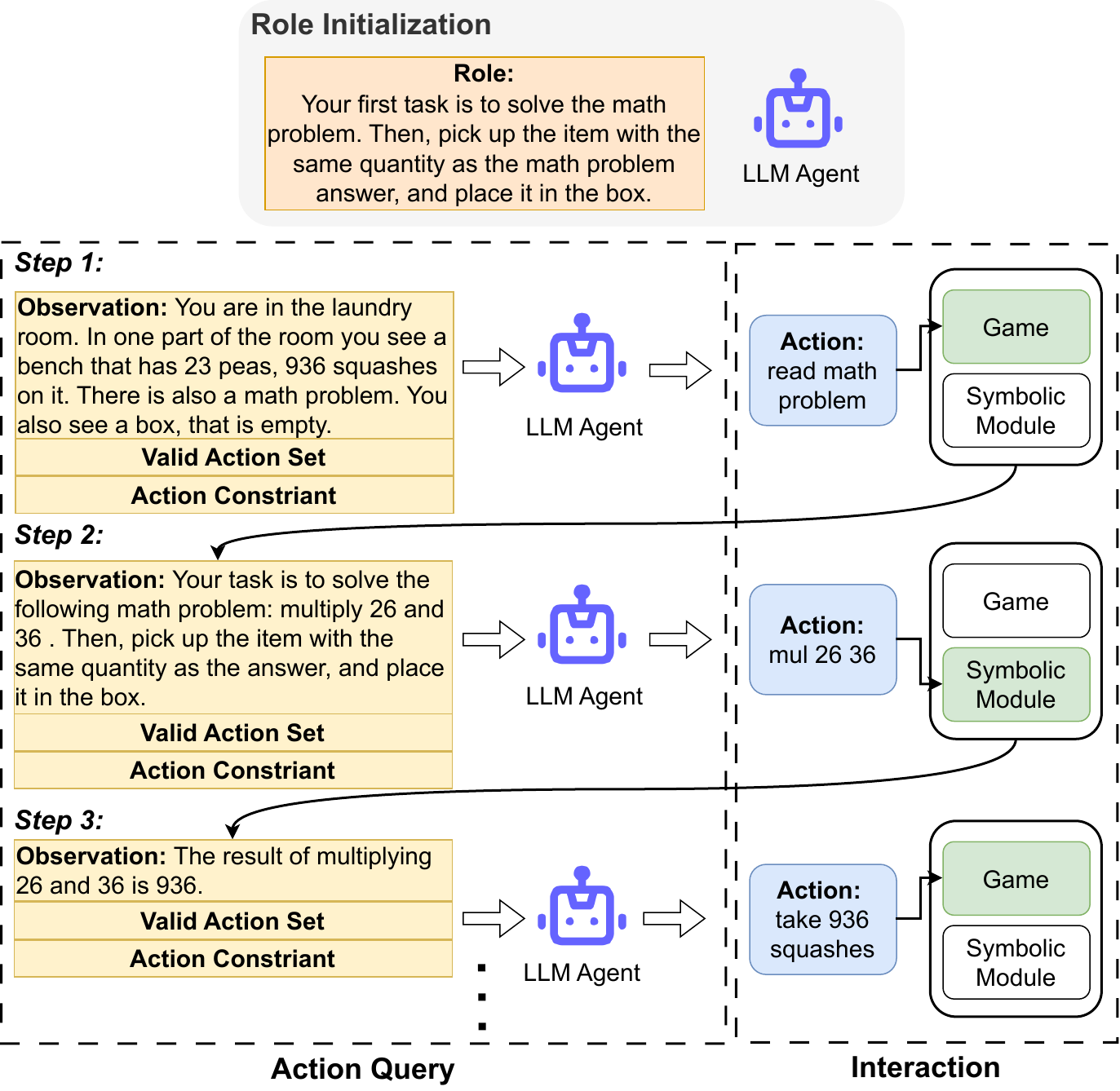} % Reduce the figure size so that it is slightly narrower than the column.
\caption{An overview of how an LLM agent plays text-based games with external symbolic modules. The following procedural steps are involved in utilizing the LLM agent for engaging in a text-based game. Initially, the LLM agent is provided with a role initialization prompt. The first observation received by the LLM agent comes from the text game environment. As depicted in the diagram, the selection of actions, determined by the LLM's reasoning, activates the symbolic module. Subsequently, the symbolic module provides output, including observations related to the module. Then the next action chosen by the LLM agent is influenced by the outcome from the symbolic module. This process is executed repeatedly until the end of the game.}
\label{ChatGPT-Steps}
\end{figure}

\section{Methodology}
% \vspace{-0.1cm}
%This section outlines the proposed methodology for employing LLMs\footnote{We utilize LLMs from OpenAI: https://chat.openai.com/.} to engage in text-based games by leveraging symbolic modules in a zero-shot manner. 
We introduce an LLM agent, namely a language agent, for employing LLMs\footnote{We utilize LLMs from OpenAI: https://chat.openai.com/.} to engage in text-based games by leveraging symbolic modules in a zero-shot manner. We begin with an overview of playing games using symbolic modules, followed by a detailed description of the key design features of our language agent, including its prompting mechanism.
%In this research, we utilize the LLM known as ChatGPT\footnote{https://chat.openai.com/}, which has been trained using a technique called reinforcement learning from human feedback (RLHF). This particular version of ChatGPT has been developed based on the insights gained from training earlier versions of the GPT model. 
%We start with an overview of the symbolic modules utilized by LLMs, and then the entire procedure with prompting. %and the prompt engineering for the experiments are introduced.
% This section presents the proposed method, how to utilise LLM to play text-based game using symbolic modules in a zero-shot manner. The LLM we use in this paper is ChatGPT\footnote{https://chat.openai.com/}, which was trained with reinforcement learning from human feedback (RLHF) from a previous GPT version. Starting with an overview of the symbolic modules utilized by LLM. Subsequently, the entire procedure is introduced and the prompt engineering for the experiments.

% The training data consists of real and generated data as the following.
% \paragraph{Real data}  Real data is sampled from the Golden Trajectories in \citep{bc_symbolic} (as shown in Table~\ref{tab:real_data})

%ChatGPT Plays TGs with Symbolic Task}
% \added[id=yudi]{add short name for TGs with symbolic tasks}
% \paragraph{Generate data by LLM}
% \paragraph{Play TG by ChatGPT}
\subsection{Playing Games with Symbolic Tasks}
We describe the process of playing games that involve symbolic tasks, using the LLM agent in conjunction with external symbolic modules.

\paragraph{Symbolic Modules.}
Symbolic modules play a crucial role in maximizing the reasoning capabilities of LLMs. For example, as shown in Figure \ref{ChatGPT-Steps}, consider a scenario where a mathematical problem is presented, and a calculator is available. In such cases, the LLM's reasoning can effectively use the calculator to complete the task in a zero-shot manner. Furthermore, symbolic modules are adept at their functions, as employing an external tool like a calculator is considered an action in itself.

The scenarios include four distinct symbolic modules: the Calculation Module, Sorting Module, Knowledge Base Module, and Navigation Module. Table \ref{tab:task&module} shows examples of how these symbolic modules are utilized. The observation produced by a symbolic module indicates the current state of the game, while the action selected by the LLM agent serves as the input. Additionally, the Navigation Module requires the previous observation as input to accurately determine the player's current position. For instance, in a mathematical task, the LLM agent may select a computational action such as ``multiply 8 by 7'' (mul 8 7). This action triggers the symbolic module to calculate the product, and the resulting observation, ``Multiplying 8 and 7 results in 56,'' is then returned.

The process of engaging in text-based games with LLMs involves multiple stages. The specifics of these steps are detailed in Figure \ref{ChatGPT-Steps}. As mentioned earlier, the comprehensive environment, comprising both the symbolic modules and the text-based game environment, presents the LLM agent with a list of allowable actions. Upon receiving an observation, the LLM agent uses its symbolic reasoning to select an action from this list. If the chosen action involves the symbolic module, the module provides the next observation; otherwise, the text-based game environment supplies the subsequent observation.

%\subsection{Prompt Engineering}
\subsection{LLM as the Neurosymbolic Reasoner}
% The experiment does not require any training, and the model's weights are frozen as it solely relies on zero-shot in-context learning. 
We investigate whether the accumulated world knowledge of LLMs can aid in making accurate decisions for downstream symbolic tasks. To ground LLMs in text-based games, we employ a prompting approach, which eliminates the need for costly additional training. Therefore, we construct prompts in a way that incorporates external context, enabling the LLM agent to generate reasonable actions. %There are three crucial components of prompts. 

%We describe the role of the agent, incorporate the observation, valid actions, and the constraints of executing the action in the prompt,  since it is not easy for LLM agents to understand the underlying rules of the environment through interacting with the game environments.
We describe the role of the agent, incorporating the observation, valid actions, and the constraints of executing the action in the prompt, as it is not easy for the LLM agent to understand the underlying rules of the environment through interacting with game environments. The key components of our approach include:
%We briefly introduce the modules as follows:
\begin{itemize}
    \item \textbf{Role initialization:} We initialize the agent by providing them with task descriptions and action constraints.
    \item \textbf{Action Query:} This step is repeated at each timestep. We prompt the LLM agent with the current observation, inventory state, valid action set, and a question.
    \item \textbf{Answer by the LLM agent:} The LLM agent chooses an action from the valid action set to complete the task.
\end{itemize}
%1) \textbf{Role initialization}. We initialize the agent by giving them some task descriptions and act constraints. 

%2) \textbf{Action Query}. This step would be repeated at each timestep, we prompt the LLM agent by the current observation, inventory state, the valid action set, as well as the question. 

%3) \textbf{Answer by the LLM agent}. LLM agent will choose an action from the valid actions set to complete the task.
%ChatGPT will call the chat completion function to give the chosen action from the valid actions set to finish the task.

\begin{table}[t]
    \centering
    % \begin{tabularx}{0.5\textwidth}{X|X}
    \begin{tabularx}{\linewidth}{>{\hsize=.2\hsize}X|X}
    \hline
    Role Initialization & \noindent You are a robot. \taskdesc \n \ You are required to choose action from the valid action set to complete the task step by step.\n \ To take action, respond with an action in the valid action set. \n \\ \hline
     Action Query & \noindent \obs  \n  \invstate \n \ Your current score is: \score \n \ The valid action set contains: \validAction.\n \ Please choose one action from the valid action set to finish the task step by step.\n \ 
     Do NOT respond with any other text, and you cannot decline to take an action. \\ \hline
    \end{tabularx}
    \caption{The prompting format for role initialization and action query for each time step. \taskdesc \ is the task description. \obs \ is the current observation. \invstate \ describes the items in your inventory. \score \ is the obtained reward. \validAction \ is a set of valid actions at the current time step.}
    \label{tab:prompt_template}
\end{table}

\paragraph{Role Initialization.}
We initialize the role and provide instructions for a functional agent assigned to a task. This process informs the agent about its role, the task description, and the actions it can take, along with their explanations and constraints. These actions are necessary for interacting with text-based games or calling the symbolic module. The agent is instructed to choose from a valid set of actions, such as reading the map, getting paths to specific locations, and recalling the task. Additionally, the agent is advised to utilize the external symbolic module and to avoid unnecessary actions during the task.
%\paragraph{Environment and Symbolic Subtask} We follow \citep{bc_symbolic} to evaluate our method in four games with corresponding symbolic subtasks. We refer reader to Table~\ref{tab:envs} for the details.

% \paragraph{Reflection of symbolic Module} 

%\paragraph{Action Query.} At each time step, we inform the LLM agent of the current game state, as outlined in Table~\ref{tab:prompt_template}, . This includes the player's observation, the state of the inventory, the reward, and the valid action set. Moreover, the inventory state pertains to the current goods of the agent. In the context of mathematical tasks, the inventory state may consist of a mathematical problem. Conversely, in the MapReader task, the inventory state may comprise a map. Additionally, the inventory state can encompass tangible objects, such as toothpaste or a quantity of 18 avocados, which have been acquired by the agent within the environment. The LLM agent is then tasked with selecting one action from the valid action set to continue with the task. It is important to note that the LLM agent is not permitted to decline or offer any text beyond the prescribed response. We also reduce the number of valid actions provided by the symbolic module. 

\begin{table}[t]
    \centering
    \begin{tabularx}{\linewidth}{l|X}
    \hline
    Task & Constrained Prompts \\ \hline
    Arithmetic & There are some rules for choosing action: \n \  
                      1) If you do not see the items that meet your  requirements, please choose `look around'.\n \
                      2) If you want to put something in the box, please first take it and then put it in box.\n \
                      3) For example, if you want to put 20 apples  in the box, you should first choose `take  20 apples' and then choose `put 20 apples  in box'.\n \
                      4) The next action of `take math problem' is `read math problem'.\n \ 
                      5) However, please never choose `put math  problem in box' as action.\n \
    \\ \hline
    MapReader &
     1) At the beginning choose `read map' to get the unknown surrounding layout.\n \
     2) After that, if you do not know how to get to SOMEPLACE, you can choose `next step to SOMEPLACE' to get the path to SOMEPLACE.\n \
     3) To choose the action, `task', you can recall  your task.\n \
     4) Do NOT go to anywhere that is unnecessary for finishing the task.\n \
     \\ \hline
     Sorting & To sort the items one by one, please follow the instruction:\n \
     1) choose `sort ascending' or `sort descending' to know which one should be sort next.\n \
     2) take the items.\n \
     3) put the items in box.\n \
     \\ \hline
     TWC & 
     1) When you take the item, you will get  positive score.\n \
     2) When you put the item in the right place, you will get higher positive score.  Otherwise you get 0.\n \ 3) You are supposed to get as much score as possible.\n \
     \\ \hline
     
    \end{tabularx}
    \caption{The prompting format for adding constraints on the actions of an agent.}
    \label{tab:prompt_constrained}
\end{table}

\paragraph{Action Query.} At each timestep, we inform the LLM agent of the current game state, as outlined in Table~\ref{tab:prompt_template}. This information includes the player's observation, the state of the inventory, the reward, and the valid action set. The inventory state refers to the current possessions of the agent. For instance, in mathematical tasks, the inventory state may consist of a mathematical problem, while in the MapReader task, it could include a map. Additionally, the inventory state can encompass tangible objects, such as toothpaste or a quantity of 18 avocados, acquired by the agent within the environment. The LLM agent is then tasked with selecting one action from the valid action set to continue with the task. It is important to note that the LLM agent is not allowed to decline or provide any text beyond the prescribed response. We also limit the number of valid actions provided by the symbolic module.
In addition, it is essential to develop appropriate prompts that effectively restrict the agent's actions according to the information provided in Table \ref{tab:prompt_constrained}. It is not feasible for the agent to acquire knowledge and infer the rules within trajectories solely through its interaction with the environment. In all tasks, there is typically a specific order of events, where the object is first taken and then placed in a designated location. This strategy is adopted to prevent scenarios where the object is placed before it is acquired, which would be considered unacceptable in the given context.

\section{Experiments}

%In this section, we proceed to evaluate our methodology in four text-based games that involve different symbolic tasks. This evaluation process follows in line with the evaluation framework outlined in the research conducted by \citep{bc_symbolic}. The experimental prompts are presented in Table \ref{tab:prompt_template} and \ref{tab:prompt_constrained}, and the process by which ChatGPT engages in TGs is illustrated in Figure \ref{ChatGPT Steps}.
We demonstrate the potential of LLMs in serving as neurosymbolic reasoners for text-based games. In particular, we present experimental results on four text-based games that involve different symbolic tasks. In these tasks, we observe that LLMs can effectively function as symbolic reasoners.

\subsection{Setup}
%This evaluation process follows in line with the evaluation framework outlined in the research conducted by \citep{bc_symbolic}. 
%The experimental prompts are presented in Table \ref{tab:prompt_template} and \ref{tab:prompt_constrained}, and the process by which ChatGPT engages in TGs is illustrated in Figure \ref{ChatGPT Steps}.
We follow the evaluation framework and game environments in \citet{bc_symbolic}. These games are developed using the TextWorldExpress game engine \cite{Jansen2022TextWorldExpressST}. For our LLM agent, we use GPT-3.5-turbo. The LLM agent can interact with game environments and symbolic modules. The task descriptions and examples of how the symbolic modules are called are provided in Table \ref{tab:task&module}. The evaluation includes four text-based games involving symbolic tasks.
%, namely MapReader, Math, Sort, and Text World Common Sense \cite{Murugesan2020TextbasedRA}. These games are developed using the TextWorldExpress game engine \cite{Jansen2022TextWorldExpressST}. In this paper, 
Each task is divided into ``Train'', ``Dev'', and ``Test'' sets. All evaluations are conducted on the ``Test'' set. 

%\paragraph{Metric.} 
The evaluation metric is based on two factors: the average score achieved at the end of each game, and the average number of steps taken within a single episode.

\begin{table*}[t]
\centering
\begin{tabular}{lcccccccccc}
\hline & \multicolumn{4}{c}{ DRRN } & \multicolumn{4}{c}{ Behavior Cloned Transformer } & \multicolumn{2}{c}{ LLM Agent } \\
& \multicolumn{2}{c}{ Baseline } & \multicolumn{2}{c}{ +symbolic module } & \multicolumn{2}{c}{ Baseline } & \multicolumn{2}{c}{ +symbolic module } & \multicolumn{2}{c}{ ~ } \\\cline {2-11} 
%\hline & \multicolumn{4}{c}{ DRRN } & \multicolumn{4}{c}{ Behavior Cloned Transformer } & \multicolumn{2}{c}{ ChatGPT } \\
%& \multicolumn{2}{c}{ Baseline } & \multicolumn{2}{c}{ NeuroSymbolic } & \multicolumn{2}{c}{ Baseline } & \multicolumn{2}{c}{ NeuroSymbolic } & \multicolumn{2}{c}{ NeuroSymbolic } \\\cline {2-11} 

Benchmark & Score & Steps & Score & Steps & Score & Steps & Score & Steps & Score & Steps\\
\hline 
Arithmetic & 0.17 & 10 & 0.14 & 7 & 0.56 & 5 & 1.00 & 5 & 1.00 & 4 \\
MapReader & 0.02 & 50 & 0.02 & 50 & 0.71 & 27 & 1.00 & 10 & 0.86 & 15\\
Sorting & 0.03 & 21 & 0.03 & 18 & 0.72 & 7 & 0.98 & 8 & 0.71 & 7 \\
TWC & 0.57 & 27 & 0.37 & 34 & 0.90 & 6 & 0.97 & 3 & 0.94 & 4  \\
\hline Average & 0.20 & 27 & 0.14 & 27 & 0.72 & 11 & 0.99 & 7 & 0.88 & 7  \\
\hline
\end{tabular}
\caption{The average performance of the model across a set of 100 games in the unseen test set. ``+symbolic module" indicates the utilization of symbolic modules within the action space of the models.}
\label{tab:chatgpt_result_with_baseline}
\end{table*}

\subsection{Environments}
%\label{appendix:tasks}
%The subsequent paragraphs provide an in-depth description of the four distinct tasks that were executed during our experimental analysis. 
We use four text-based game benchmark environments \cite{bc_symbolic}:

\paragraph{Arithmetic.} The task at hand involves a mathematical component, wherein an agent is required to read and solve a mathematical problem. This process determines the specific object from a given set of objects that they should select and place. The arithmetic game includes a calculator module equipped with the capability to perform basic mathematical operations, including addition, subtraction, multiplication, and division.
%The task at hand involves a mathematical component, wherein agents are required to read and solve a mathematical problem in order to determine the specific object from a given set of objects that they should select and place. An example instance of a task assigned to an agent could involve collecting of an item that corresponds to the sum of 34 and 43, and thereafter placing such item into the specified box. The arithmetic game incorporates a calculator module that is equipped with the capability to do basic mathematical operations, including addition, subtraction, multiplication, and division. As a result, the symbolic module can provide the solution to the addition of 34 and 43, which is 77. Once the agent acquires an object with a quantity of 77, such as 77 eggplants, the task will be recognized as accomplished. 

% A math-themed task, where agents must read and solve a math problem in order to know which object from a set of objects to pickand-place. An example problem is “take the bundle of objects that is equal to 3 multiplied by 6, and place them in the answer box”, where the agent must complete the task by choosing 18 apples. Distractor objects are populated with quantities that correspond to performing the arithmetic incorrectly (e.g. 3 oranges, corresponding to subtracting 3 from 6). We pair the arithmetic game with a calculator module capable of performing addition, subtraction, multiplication, and division.

\paragraph{MapReader.} A pick-and-place game with a navigation theme, similar to the Coin Collector game \cite{yuan2018counting}. The agent is equipped with a map that may be exploited to optimize route planning. The map provides information on the connections between rooms, such as the lounge connecting to the cookery and supermarket. The navigation symbolic module has the capability to extract location information from the observation space. This includes specific information relating to the present location and geographical features leading to the intended destination. For instance, the instructions sent to the agent might indicate that in order to get from the cooking area to the recreation zone, one must pass through the bar, steam room, library, and finally reach the recreation zone.

\paragraph{Sorting.} This game involves an agent initially situated in a room containing a variable number of objects, ranging from three to five. The agent's task is to sequentially place these objects into a designated box, adhering to a specific sorting criterion based on increasing quantity. In this game, units related to volume, mass, or length are used, as exemplified by items such as 25g of oak, 12ml of marble, and 6cm of cedar. The sorting game includes a module capable of extracting information from the observation space. This module is specifically designed to identify items that include quantities and can arrange these objects in either ascending or descending order, following the user's instructions.

\paragraph{Text World Common Sense (TWC).} The challenges provided in this game serve as a baseline for evaluating common sense reasoning abilities \cite{murugesan2021text}. In this game, agents are required to gather objects from their surroundings, such as a clean brown shirt, and subsequently place these objects in their appropriate and commonly recognized locations, like a wardrobe. The incorporation of a symbolic module within this game enables agents to engage in knowledge-based queries. For instance, it allows them to deduce that a clean brown shirt is typically found in a wardrobe.
% A benchmark common sense reasoning task where agents must collect objects from the environment (e.g. dirty socks), and place those objects in their canonical common sense locations (e.g. washing machine). The symbolic module for this game allows agents to query a knowledge base of (subject, relation, object) triples (e.g. (cushion, hasCanonicalLocation, sofa)).

%The agent begins its journey from a randomly selected starting point, such as the cookery, and is given a designated destination, such as the recreation zone. The agent must successfully get to the designated destination, retrieve a coin, then retrace its route back to the initial point of departure, and finally deposit the aforementioned coin into a box located at the specified location. 

% A navigation-themed pick-and-place game similar to Coin Collector . An agent starts in a random location (e.g. the kitchen), and is provided with a target location (e.g. the garage). The agent must navigate to the target location, pick up a coin, then return to the starting location and place it in a box. The agent is further provided with a map that can be used for efficient route planning. The navigation symbolic module
% paired with this environment scrapes the observation space for location information (e.g. you are currently in the kitchen), and both complete (e.g. the map) or partial (e.g. to the north you see the hallway) spatial connection information.

\subsection{Baselines}
We also compare our LLM agent with two baselines, namely the Deep Reinforcement Relevance Network (DRRN) \cite{he2016deep} and the T5-based Behavior Cloned Transformer \cite{raffel2020exploring, bc_symbolic}, as follows:
\begin{itemize}

 %  \item \textbf{DRRN} The primary notion of DRRN is built on the Q-learning. The candidate action with the greatest anticipated Q-value will be chosen as the next action based on the current observation. DRRN employs a Deep Q-Network \cite{mnih2013playing} to estimate the Q-value for each pair of observation-action that is encoded by the method in \cite{cho2014properties}. \citet{xu2020deep} notes that the DRRN is a rapid and strong reinforcement learning baseline that is frequently utilized to produce near state-of-the-art performance in a range of text games.

\item \textbf{DRRN:} The primary concept of the DRRN is based on Q-learning. The candidate action with the highest anticipated Q-value is chosen as the next action, based on the current observation. The DRRN employs a Deep Q-Network \cite{mnih2013playing} to estimate the Q-value for each observation-action pair. \citet{xu2020deep} note that the DRRN is a fast and robust reinforcement learning baseline, frequently used to produce near state-of-the-art performance in a variety of text-based games.

  % \item \textbf{Behavior Cloned Transformer} In recent years, the Behavior Cloned technique \cite{torabi2018behavioral} has garnered significant attention due to its notable performance. This method adopts an imitation learning approach that conceptualizes reinforcement learning as a sequence-to-sequence problem which is equivalent to the Decision Transformer \cite{chen2021decision}. It accomplishes this by forecasting the subsequent action based on a sequence of preceding observations. The baseline is referenced by \cite{bc_symbolic} and aligns with the approach outlined in \cite{ammanabrolu2020motivate}, where the model input at time step $t$ encompasses the task description, present state observation, previous action, and previous state observation. Symbolic modules are utilized in the demonstration, specifically employing gold trajectories.
 \item \textbf{Behavior Cloned Transformer:} This method adopts an imitation learning approach, conceptualizing reinforcement learning as a sequence-to-sequence problem, similar to the Decision Transformer \cite{chen2021decision}. It predicts the subsequent action based on a sequence of previous observations. This baseline aligns with the approach described in \citet{ammanabrolu2020motivate}, where the model input at timestep $t$ includes the task description, current state observation, previous action, and previous state observation. Symbolic modules are utilized in the demonstrations, specifically employing gold trajectories.
\end{itemize}

Following \citet{bc_symbolic}, both baseline models include two variants: one with symbolic modules and one without. When using symbolic modules, we inject actions from these modules into the action space of each game for the baseline models.

\subsection{Results}

%Based on the baseline results presented in Table \ref{tab:chatgpt_result_with_baseline}, it can be observed that the utilization of the symbolic module, when combined with the LLM agent, yields a favorable average performance in comparison to alternative baseline approaches. When comparing the outcomes of the behavior cloned transformer with symbolic module to those of the LLM agent, it can be observed that the performance of the LLM agent is slightly lower. Additionally, the process of interacting with the game environment exhibits a similar level of competency. In addition, the LLM agent does not undergo extensive training with a large volume of expert data, as is typically employed in training behavior cloning transformer models. Consequently, this approach conserves significant training resources. 

Based on the results presented in Table \ref{tab:chatgpt_result_with_baseline}, it is evident that the use of the symbolic module in conjunction with the LLM agent yields a favorable average performance compared to other baseline approaches. When comparing the outcomes of the Behavior Cloned Transformer with a symbolic module to those of the LLM agent, the performance of the LLM agent is observed to be slightly lower. However, the LLM agent demonstrates a similar level of competency in interacting with the game environment. Furthermore, unlike the Behavior Cloned Transformer models, the LLM agent does not require extensive training with a large volume of expert data. As a result, this approach saves significant training resources.

Table \ref{tab:chatgpt_result} demonstrates that the LLM agent possesses a robust capacity for reasoning, enabling effective handling of tasks involving symbolic tasks. It shows exceptional performance, particularly in mathematics. In the MapReader benchmark, the agent achieves commendable scores, though it requires a considerable number of steps to complete the task. This inefficiency is mainly due to the agent's tendency to forget the route obtained from the symbolic module, leading to the risk of reaching incorrect locations and necessitating repeated route queries. The complexity of map logic, which involves determining one's current location and desired destination, adds to the probabilistic nature of this task. In contrast, the Sorting task reveals suboptimal performance, as the LLM agent's understanding of sorting logic is not fully developed. This issue is largely attributed to the agent's limited memory capacity, hindering its ability to remember the ascending order of all objects. 
%The details of these failure cases will be further evaluated in the following section.

%In Table \ref{tab:chatgpt_result_ablation}, an ablation study is performed to compare the performance of the model with constrained prompts to the model without constrained prompts. It shows that when the LLM agent is provided with the prompts outlined in Table \ref{tab:prompt_constrained}, there is an improvement in performance across all tasks. Additionally, there is a reduction in the average number of steps required to communicate with the game environment. This serves as evidence to support the effectiveness of our constrained prompts in these tasks. Additionally, we provide the experimental results using GPT4, as shown in Table~\ref{tab:gpt4}. It significantly performs better than the GPT-3.5 agent in MapReader and Sorting task, while demonstrating worse performance on TWC.

In Table \ref{tab:chatgpt_result_ablation}, it compares the performance of the model with constrained prompts to that of the model without constrained prompts. The results indicate that when the LLM agent is provided with the prompts outlined in Table \ref{tab:prompt_constrained}, there is an improvement in performance across all tasks. Additionally, a reduction in the average number of steps required to interact with the game environment is observed. This demonstrates the effectiveness of our constrained prompts in these tasks. Furthermore, experimental results using GPT-4, as shown in Table~\ref{tab:gpt4}, reveal that it significantly outperforms the GPT-3.5 agent in the MapReader and Sorting tasks, while showing weaker performance in the TWC task.

\paragraph{Discussion.} Our results demonstrate that the incorporation of external symbolic modules by the LLM agent leads to enhanced average accuracy compared to other baselines. This capability is achieved by leveraging the underlying patterns present in the training data. Instead of relying on symbolic thinking or explicit rules, this approach acquires knowledge by recognizing patterns and associations from the extensive corpus of text to which it has been exposed during its training phase, as exemplified by GPT-3.5 and GPT-4 \cite{OpenAI_GPT4_2023}. Although the LLM agent has the capability to connect with a symbolic module for specific tasks, it still exhibits uncertainty and is prone to making mistakes.

\begin{table}[t]
\renewcommand{\arraystretch}{1.1} % Default value: 1
\setlength{\tabcolsep}{5.5pt} % Default value: 6pt
\small
\centering
\begin{tabular}{lllllll}
\hline
\multirow{2}{*}{Task} & \multicolumn{2}{c}{Train} & \multicolumn{2}{c}{Dev}  & \multicolumn{2}{c}{Test} \\ \cline{2-7}
                      & \multicolumn{1}{c}{Score} & \multicolumn{1}{c}{Steps} & \multicolumn{1}{c}{Score} & \multicolumn{1}{c}{Steps}  & \multicolumn{1}{c}{Score} & \multicolumn{1}{c}{Steps} \\ \hline
 
% MapReader (w/o Subgoal)  & 0.64 & 36.35 & 0.65 & 33.97 & 0.64 & 34.72 \\   
Arithmetic  & 1.00  & 3  & 0.95 & 4 & 1.00 & 4  \\ 
MapReader & 0.84 & 15 & 0.84 & 14 & 0.86 & 15 \\  
Sorting  & 0.70  & 7 & 0.63 & 6 & 0.71 & 7  \\
TWC         & 0.93 & 4 & 0.835 & 5 & 0.94 & 4    \\ 
\hline Average & 0.87 & 7 & 0.81 & 7 & 0.88 & 7 \\ \hline
\end{tabular}
% \end{small}
%\vspace{-0.3cm}
\caption{The performance of the LLM agent on different sets of the game, including ``Train'', ``Dev'', and ``Test''. The scores are subjected to normalization, resulting in values ranging from 0 to 1, with higher values indicating greater performance. On the other hand, the steps quantify the number of actions taken by an agent inside the environment, with lower values indicating more efficient behavior. }

\label{tab:chatgpt_result}
\end{table}

\begin{table}[t]
\renewcommand{\arraystretch}{1.1} % Default value: 1
\setlength{\tabcolsep}{5.5pt} % Default value: 6pt
%\small
\centering
\begin{tabular}{lllll}
\hline
\multirow{2}{*}{Task} & \multicolumn{2}{c}{w/ Constraints} & \multicolumn{2}{c}{w/o Constrains}   \\ \cline{2-5}
                      & \multicolumn{1}{c}{Score} & \multicolumn{1}{c}{Steps} & \multicolumn{1}{c}{Score} & \multicolumn{1}{c}{Steps}  \\ \hline
 
% MapReader (w/o Subgoal)  & 0.64 & 36.35 & 0.65 & 33.97 & 0.64 & 34.72 \\   
Arithmetic  & 1.00  & \ \ \  4 & 0.96 & \ \ \ \ \ 3   \\ 
MapReader & 0.86 & \ \ \ 15 & 0.64 & \ \ \ \ \ 12  \\  
Sorting  & 0.71  & \ \ \  7 & 0.35 & \ \ \ \ \ 10   \\
TWC   & 0.94 & \ \ \  4 & 0.73 & \ \ \ \ \ 7    \\ 
\hline
Average & 0.88 & \ \ \  7 & 0.67 & \ \ \ \ \ 8 \\ \hline
\end{tabular}
% \end{small}
%\vspace{-0.3cm}
\caption{The performance of the LLM agent with and without constrained prompts on the ``Test'' set. The constrained prompts are shown in Table \ref{tab:prompt_constrained}.}
\label{tab:chatgpt_result_ablation}
\end{table}

\begin{table}[t]
% \renewcommand{\arraystretch}{1.1} % Default value: 1
% \setlength{\tabcolsep}{5.5pt} % Default value: 6pt
%\small
\centering
\begin{tabular}{lllll}
\hline
\multirow{2}{*}{Task} & \multicolumn{2}{c}{w/ GPT-3.5} & \multicolumn{2}{c}{w/ GPT-4}   \\ \cline{2-5}
                      & \multicolumn{1}{c}{Score} & \multicolumn{1}{c}{Steps} & \multicolumn{1}{c}{Score} & \multicolumn{1}{c}{Steps}  \\ \hline
  
% MapReader (w/o Subgoal)  & 0.64 & 36.35 & 0.65 & 33.97 & 0.64 & 34.72 \\   
Arithmetic  & 1.00  & \ \ \  4 & 1.00 & \ \ \ \ \ 4   \\ 
MapReader & 0.86 & \ \ \ 15 & 0.99 & \ \ \ \ \ 7  \\ 
Sorting  & 0.71  & \ \ \  7 & 0.93 & \ \ \ \ \ 8   \\
TWC   & 0.94 & \ \ \  4 & 0.71 & \ \ \ \ \ 16    \\ 
\hline
Average & 0.88 & \ \ \  7 & 0.91 & \ \ \ \ \ 8 \\ \hline
\end{tabular}
% \end{small}
\vspace{-0.06in}
% \vspace{-0.3cm}
\caption{The performance of the LLM agent using GPT-3.5 and GPT-4 on the ``Test'' set. }
\vspace{-0.10in}
\label{tab:gpt4}
\end{table}

\section{Conclusion}
%In summary, this research demonstrates the application of Large Language Models (LLMs) in the training of text-based games that involve symbolic tasks
%This paper has demonstrated the application of Large Language Models (LLMs) in complex text-based games that involve symbolic tasks, thereby diminishing the need for data generated by humans. By employing a zero-shot learning methodology, we instruct ChatGPT to effectively engage in text-based games. The efficacy of our approach utilizing ChatGPT has demonstrated superiority over alternative benchmarks, thereby underscoring the potential of LLMs in enhancing the training procedures of text-based games. In alternative terms, it can be proposed that Large Language Models have characteristics of Neurosymbolic Reasoners due to their adeptness in effectively executing symbolic tasks through the utilization of logical reasoning.
This paper has demonstrated the effective application of Large Language Models (LLMs) in complex text-based games involving symbolic tasks. Utilizing a prompting approach, we have guided the LLM agent to efficiently engage with symbolic modules within these games. The efficacy of our method, leveraging LLMs, has shown superior performance compared to alternative benchmarks, highlighting the potential of LLMs to enhance training procedures in text-based games. Consequently, it can be posited that Large Language Models can be considered as Neurosymbolic Reasoners, possessing significant potential for performing symbolic tasks in real-world applications.

\subsection{Limitations}
%The inclusion of additional prompts would provide greater control over the actions of the agent, which would be particularly beneficial in tasks such as sorting, where it is desirable to have essential information pre-provided. Taking into consideration these limits has the potential to enhance the system's performance. In order to make future progress, it is imperative to expand the model's utilisation to more complex domains that go beyond the context of straightforward text-based games. The integration of complicated symbolic modules would be required in order to address the complexities of various scenarios, hence facilitating a more proficient problem-solving methodology.
The addition of more detailed prompts could offer greater control over the actions of the LLM agent. This would be particularly beneficial in tasks like Sorting, where providing essential information beforehand is advantageous. Acknowledging and addressing these limitations could significantly enhance the system's performance. For future progress, it is crucial to extend the model's application to more complex domains, going beyond the scope of straightforward text-based games. Integrating more sophisticated symbolic modules would be necessary to tackle the complexities of diverse scenarios, thereby facilitating a more efficient problem-solving approach.

% The primary limitation of the current architecture lies in its absolute dependence on zero-shot capabilities for ChatGPT. The incorporation of demonstrative game trajectories in the integration of few-shot learning has the potential to enhance performance and improve the robustness of the model.

% The primary constraint inherent in our current framework is the exclusive reliance on zero-shot capabilities for our ChatGPT system. The integration of few-shot learning into the framework has the potential to enhance performance. For instance, a limited number of game trajectories might be provided as illustrations, and given the consistent underlying logic of the game, employing few-shot learning techniques may prove effective in enhancing performance. Additionally, the inclusion of additional prompts could serve to exert greater control over the actions undertaken by the agent. In the context of a sorting task, if the agent is unable to recall the ascending order of the objects or the quantity of objects being sorted, it would be beneficial to include this information in the prompts prior to the model generating the action. 

% {1 current performance is not perfect. 2 It is better to apply llm agent to more complex environments. Text-based games are still not very complex. 3 more symbolic modules}

\bibliography{aaai24}

\end{document}